\documentclass[letterpaper, preprint, paper,11pt]{AAS}	

\usepackage{bm}
\usepackage{amsmath}
\usepackage{subfigure}
\usepackage[colorlinks=true, pdfstartview=FitV, linkcolor=black, citecolor= black, urlcolor= black]{hyperref}
\usepackage{overcite}
\usepackage{footnpag}			      	
\usepackage{float}
\PaperNumber{26-893}

\begin{document}

\title{Tightly-Coupled Estimation and Guidance for Robust Low-Thrust Rendezvous via Adaptive Homotopy}

\author{Batu Candan\thanks{Ph.D. Candidate, Department of Aerospace Engineering, Iowa State University, Ames, IA 50011.},
\ and Simone Servadio\thanks{Assistant Professor, Department of Aerospace Engineering, Iowa State University, Ames, IA 50011.}
}

\maketitle{}

\begin{abstract}
Minimum-fuel low-thrust rendezvous guidance yields bang-bang control structures that are highly sensitive to estimation errors, non-Gaussian sensor anomalies, and solver regularization, making purely aggressive closed-loop execution brittle for uncooperative proximity operations. This paper proposes a tightly-coupled estimation and guidance architecture in which navigation confidence directly modulates the homotopy parameter of a receding-horizon indirect optimal control solver. Relative motion is modeled in the Clohessy--Wiltshire frame, and the translational state is estimated with a linear Kalman filter augmented by a measurement-side Multiple Tuning Factors (MTF) covariance inflation mechanism that suppresses suspicious innovation directions. A composite innovation score, constructed from the normalized innovation and the MTF activity, is then mapped online to the homotopy continuation parameter, allowing the controller to relax toward a smoother and more conservative regime when confidence degrades and to recover a more fuel-efficient bang-bang structure as sensing improves. Numerical results under severe measurement degradation show that fixed bang-bang guidance remains brittle, as both the plain-KF and MTF-KF fixed-$\epsilon$ controllers exhibit large terminal miss distances. In contrast, the proposed MTF-adaptive homotopy controller reduces the terminal miss by roughly two orders of magnitude, bringing the final error from the scale of hundreds of meters down to the sub-meter level, while requiring only a moderate increase in control effort relative to the open-loop fuel-optimal benchmark. A comparison study further indicates that adaptive homotopy is the dominant robustness mechanism, while MTF provides an additional improvement in both accuracy and efficiency within the adaptive architecture. The receding-horizon implementation also exhibited consistently fast and reliable solution times, supporting the practical online viability of the proposed method for the tested scenario.
\end{abstract}

\section{Introduction}

Autonomous rendezvous with uncooperative objects is a key enabling capability for active debris removal, on-orbit servicing, inspection, and future in-space assembly missions. In these settings, the target may be poorly characterized, tumbling, and observed through degraded sensing conditions, so estimation, planning, and control must operate robustly and with limited opportunity for human intervention [\citen{albee2021robustPipeline, phdthesis, serva2}]. At the same time, low-thrust propulsion is especially attractive because of its high specific impulse, fine control authority, and favorable mission-level fuel efficiency, which has motivated a broad literature on low-thrust trajectory design and spacecraft guidance [\citen{morante2021surveyLowThrust,wang2024surveyConvexGC,malyuta2021advancesTrajectory,gurfil2023constantMagnitude}]. The main difficulty is that minimum-fuel low-thrust guidance is numerically demanding. Direct transcription, convexification, and model-predictive approaches have become particularly attractive because they can accommodate path and control constraints and can be adapted to receding-horizon operation [\citen{malyuta2022convexTutorial,leomanni2014explicitMPC,bhagat2016convexGuidanceEnvisat,hofmann2021closedLoopConvex,hofmann2021rapidConvex,hofmann2025homotopicHighFidelity,pavanello2025collisionAvoidance,malyuta2023fastHomotopyDiscrete,servadio2023koopman}]. In parallel, learning-based surrogates and controllers are increasingly being explored to reduce onboard computational cost or to provide fast approximations of optimal guidance, including supervised surrogates, deep warm starts, reinforcement learning, and certifiably stable neural guidance laws [\citen{li2020deepNetworks,cheng2019multiscaleDNN,zavoli2021rlRobust,yang2026knowledgeDNN,wang2025learningStable,zhang2025neuralApproximators,wang2026ocnnSurvey}]. Taken together, recent surveys on low-thrust optimization, convex guidance, and learning for optimal control show how rapidly this design space is expanding [\citen{morante2021surveyLowThrust,wang2024surveyConvexGC,malyuta2021advancesTrajectory,wang2026ocnnSurvey}].

Indirect methods remain compelling because, once properly initialized, they can deliver highly accurate fuel-optimal solutions with a smaller decision space by exploiting Pontryagin's Minimum Principle. Their weakness, however, is equally well known as the resulting Hamiltonian boundary-value problem has a narrow convergence basin, and the bang-bang or bang-off-bang structure of minimum-fuel low-thrust control substantially increases sensitivity to the initial co-states and to the numerical regularization used during solution [\citen{taheri2016enhancedSmoothing,pan2018quadraticHomotopy,huang2024modifiedHomotopy,tafazzol2024regularizationComparison,nurre2025indirectDirect,parrish2018cislunarThesis}]. For this reason, a large body of work has focused on smoothing, continuation, and regularization strategies that embed the original fuel-optimal problem inside a family of easier auxiliary problems [\citen{taheri2016enhancedSmoothing,pan2018quadraticHomotopy,huang2024modifiedHomotopy,tafazzol2024regularizationComparison,mall2020uniformTrig,nurre2025indirectDirect}]. Related homotopic ideas have also appeared in convex and mixed-logic rendezvous optimization, where continuation improves convergence and computational reliability [\citen{malyuta2023fastHomotopyDiscrete,hofmann2025homotopicHighFidelity}]. Broader reviews of low-thrust optimization likewise emphasize the central role of homotopy and smoothing in making discontinuous minimum-fuel solutions computationally tractable [\citen{morante2021surveyLowThrust,parrish2018cislunarThesis}]. Despite this progress, the homotopy or regularization parameter is still usually treated as an optimizer-side continuation variable rather than as a closed-loop robustness actuator driven by onboard navigation. At the same time, integrated rendezvous pipelines have emphasized the importance of estimation, planning, and control under uncertainty [\citen{albee2021robustPipeline}], but the numerical aggressiveness of the low-thrust guidance law is still generally decoupled from the estimator's instantaneous confidence. This separation is especially limiting for uncooperative rendezvous, where non-Gaussian anomalies such as optical (shadowing, glare) effects, imperfect target knowledge, or unmodeled disturbances can abruptly degrade the state estimate while the guidance law continues to pursue an aggressive fuel-optimal solution.

This paper addresses that gap by proposing a tightly coupled estimation and guidance architecture for robust low-thrust rendezvous. Relative motion is modeled in the Clohessy--Wiltshire (CW) frame, and the guidance law is generated by a receding-horizon indirect solver for the low-thrust optimal control problem. To improve robustness on the estimation side, a Multiple Tuning Factors (MTF) Kalman filtering strategy is used to adapt the covariance in the presence of non-Gaussian anomalies and measurement faults [\citen{batumtf}]. The key idea is to reinterpret the normalized innovation not only as an estimation-consistency measure, but also as a control-relevant confidence signal. Specifically, the innovation is mapped online to the homotopy continuation parameter of the indirect solver. When the innovation remains small, the homotopy parameter is reduced, and the controller recovers a fuel-optimal bang-bang-like solution. When the innovation increases, the homotopy parameter is increased automatically, relaxing the problem toward a smoother and more robust energy-optimal regime. In this way, the onboard guidance law responds not only to the estimated state, but also to the reliability of that estimate. The main contributions of this work are threefold. First, we formulate a tightly coupled receding-horizon architecture in which estimator confidence directly modulates the continuation process of an indirect low-thrust solver. Second, we adapt an MTF-based covariance tuning concept to translational relative motion guidance for uncooperative rendezvous. Third, we show how innovation-adaptive homotopy enables a continuous and physically interpretable trade between fuel optimality and robustness during low-thrust close-proximity operations.

The remainder of the paper is organized as follows. Section~\ref{sec:dynamics} presents the relative-motion model, the minimum-fuel low-thrust optimal control problem, and the homotopy-regularized indirect guidance formulation. Section~\ref{sec:estimation} introduces the linear Kalman filtering framework, the measurement-side MTF adaptation, and the innovation-adaptive homotopy scheduling law. Section~\ref{sec:results} presents the numerical results, including the baseline closed-loop comparison and the adaptive-only comparison study. Finally, Section~\ref{sec:conclusion} concludes the paper and outlines directions for future work.

\section{Dynamical Model and Optimal Control} \label{sec:dynamics}
\subsection{Relative Motion Dynamics}
The relative motion of the chaser spacecraft with respect to the target is modeled in the CW frame. Let the state vector be
\begin{equation}
\bm{x} = \begin{bmatrix} x & y & z & \dot{x} & \dot{y} & \dot{z} \end{bmatrix}^T
\end{equation}
and let the control acceleration be
\begin{equation}
\bm{u} = \begin{bmatrix} u_x & u_y & u_z \end{bmatrix}^T .
\end{equation}
Assuming the target follows a circular reference orbit with mean motion $n$, the linearized relative dynamics are
\begin{equation}
\dot{\bm{x}} = A \bm{x} + B \bm{u}
\label{eq:cw_state}
\end{equation}
with
\begin{equation}
A =
\begin{bmatrix}
0 & 0 & 0 & 1 & 0 & 0 \\
0 & 0 & 0 & 0 & 1 & 0 \\
0 & 0 & 0 & 0 & 0 & 1 \\
3n^2 & 0 & 0 & 0 & 2n & 0 \\
0 & 0 & 0 & -2n & 0 & 0 \\
0 & 0 & -n^2 & 0 & 0 & 0
\end{bmatrix},
\qquad
B =
\begin{bmatrix}
0 & 0 & 0 \\
0 & 0 & 0 \\
0 & 0 & 0 \\
1 & 0 & 0 \\
0 & 1 & 0 \\
0 & 0 & 1
\end{bmatrix}.
\label{eq:cw_AB}
\end{equation}
The CW model is adopted here as a first-order translational model that is sufficiently simple for repeated onboard re-optimization while still capturing the principal relative dynamics relevant to close-proximity rendezvous.

\subsection{Minimum-Fuel Problem and Homotopy Regularization}
The guidance objective is to steer the relative state from the current estimated state to a desired terminal condition while minimizing control effort. The nominal minimum-fuel optimal control problem is written as
\begin{equation}
J = \int_{0}^{t_f} \|\bm{u}(t)\| \, dt
\label{eq:fuel_cost}
\end{equation}
subject to Eq.~\eqref{eq:cw_state} and the control bound
\begin{equation}
\|\bm{u}(t)\| \le u_{\max}.
\label{eq:control_bound}
\end{equation}

Applying Pontryagin's Minimum Principle yields the Hamiltonian
\begin{equation}
\mathcal{H} = \|\bm{u}\| + \bm{\lambda}^T (A\bm{x} + B\bm{u}),
\end{equation}
where $\bm{\lambda} = [\bm{\lambda}_r^T \ \bm{\lambda}_v^T]^T$ is the costate vector. The optimal thrust direction is
\begin{equation}
\bm{\alpha} = -\frac{\bm{\lambda}_v}{\|\bm{\lambda}_v\|},
\label{eq:alpha_dir}
\end{equation}
and the switching function is
\begin{equation}
S(t) = 1 - \|\bm{\lambda}_v(t)\|.
\label{eq:switching_function}
\end{equation}
In the fuel-optimal limit, the control is bang-bang. Because this discontinuity makes the indirect boundary-value problem difficult to solve robustly, a scalar homotopy parameter $\epsilon \ge 0$ is introduced to regularize the control magnitude. The smoothed control law is written as
\begin{equation}
\bm{u} =
\begin{cases}
0, & S > \epsilon \\
\dfrac{\epsilon - S}{2\epsilon} u_{\max}\bm{\alpha}, & -\epsilon \le S \le \epsilon \\
u_{\max}\bm{\alpha}, & S < -\epsilon .
\end{cases}
\label{eq:smoothed_control}
\end{equation}
As $\epsilon \rightarrow 0$, Eq.~\eqref{eq:smoothed_control} recovers the bang-bang fuel-optimal structure. For larger $\epsilon$, the control becomes smoother and numerically more robust. In the proposed architecture, this regularization parameter is not treated purely as an offline continuation variable; instead, it is adjusted online according to estimator confidence.

\subsection{Receding-Horizon Indirect Guidance}
At each guidance update, the current state estimate $\hat{\bm{x}}_k$ is used as the initial condition of a finite-horizon indirect optimal control problem. The corresponding costate boundary-value problem is solved numerically, and only the first segment of the resulting control is applied before the problem is re-solved at the next update. This produces a receding-horizon indirect guidance strategy that retains the structure and efficiency of the indirect formulation while allowing the controller to react to updated navigation information and changing confidence levels.

\section{Estimation and Innovation-Adaptive Homotopy}
\label{sec:estimation}
\subsection{Measurement Model and Linear Kalman Filter}
The translational state is estimated with a linear Kalman filter consistent with the CW dynamics. The discrete-time propagation model is
\begin{equation}
\bm{x}_{k+1}^{-} = \Phi \bm{x}_{k}^{+} + \Gamma \bm{u}_{k},
\label{eq:state_pred}
\end{equation}
\begin{equation}
P_{k+1}^{-} = \Phi P_k^{+} \Phi^T + Q,
\label{eq:cov_pred}
\end{equation}
where $\Phi = e^{A\Delta t}$ and $\Gamma = \int_{0}^{\Delta t} e^{A\tau} B \, d\tau$ are the exact discrete-time state-transition and control matrices over the sampling interval $\Delta t$.

The measurement model is taken as
\begin{equation}
\bm{y}_k = H \bm{x}^+_k + \bm{v}_k,
\qquad
H = \begin{bmatrix} I_{3\times3} & 0_{3\times3} \end{bmatrix},
\label{eq:meas_model}
\end{equation}
so that the relative position is measured directly. The nominal innovation and innovation covariance are
\begin{equation}
\bm{\nu}_k = \bm{y}_k - H \bm{x}_k^{-},
\label{eq:innovation}
\end{equation}
\begin{equation}
S_k = H P_k^{-} H^T + R_{\text{nominal}}.
\label{eq:innovation_cov}
\end{equation}

\subsection{Measurement-Side MTF Adaptation}
To improve robustness against non-Gaussian measurement anomalies, an MTF correction is constructed from the mismatch between the instantaneous innovation and the nominal innovation covariance, inspired by the prior works of the authors [\citen{sokencandan, candanMdpi, candan, batumtf}]. First, define
\begin{equation}
S_{k,\mathrm{raw}} = \bm{\nu}_k \bm{\nu}_k^T - S_k .
\label{eq:sraw}
\end{equation}
The MTF matrix is then formed by retaining only positive diagonal excess terms,
\begin{equation}
S_{k,\mathrm{MTF}} = \mathrm{diag}\!\left(\max\!\left(\mathrm{diag}(S_{k,\mathrm{raw}}),0\right)\right).
\label{eq:smtf}
\end{equation}
This construction identifies measurement directions whose observed innovation energy exceeds the nominal expectation and inflates only those axes. The effective measurement covariance becomes
\begin{equation}
R_{k,\mathrm{eff}} = R_{\text{nominal}} + S_{k,\mathrm{MTF}},
\label{eq:reff}
\end{equation}
and the Kalman gain is computed as
\begin{equation}
K_k = P_k^{-} H^T \left(H P_k^{-} H^T + R_{k,\mathrm{eff}}\right)^{-1}.
\label{eq:kalman_gain_mtf}
\end{equation}
The state and covariance updates are then
\begin{equation}
\bm{x}_k^{+} = \bm{x}_k^{-} + K_k \bm{\nu}_k,
\label{eq:kf_state_update}
\end{equation}
\begin{equation}
P_k^{+} = (I-K_k H)P_k^{-}(I-K_k H)^T + K_k R_{k,\mathrm{eff}} K_k^T.
\label{eq:joseph_update}
\end{equation}

In the present implementation, the MTF term is used as a measurement-side innovation-covariance inflation. Its role is to reduce the filter's trust in suspicious measurements while preserving sensitivity along well-behaved axes.

\subsection{Innovation-Based Confidence Score}
A raw normalized innovation score is computed as
\begin{equation}
s_k^{\mathrm{NIS}} = \bm{\nu}_k^T S_k^{-1} \bm{\nu}_k.
\label{eq:raw_nis}
\end{equation}
To capture not only the instantaneous innovation but also the severity of the MTF correction, a relative MTF magnitude is introduced:
\begin{equation}
\rho_k^{\mathrm{MTF}} = \frac{\mathrm{tr}(S_{k,\mathrm{MTF}})}{\mathrm{tr}(S_k)}.
\label{eq:mtf_ratio}
\end{equation}
These quantities are combined into a scalar adaptation score,
\begin{equation}
s_k^{\mathrm{comp}} = \max\!\left(s_k^{\mathrm{NIS}} - m, 0\right) + \kappa_{\mathrm{MTF}} \rho_k^{\mathrm{MTF}},
\label{eq:composite_score}
\end{equation}
where $m$ is the measurement dimension and $\kappa_{\mathrm{MTF}}$ is a tuning parameter. In practice, a first-order low-pass filter is applied to reduce sensitivity to isolated spikes:
\begin{equation}
\bar{s}_k = \rho_s \bar{s}_{k-1} + (1-\rho_s)s_k^{\mathrm{comp}} .
\label{eq:filtered_score}
\end{equation}

\subsection{Innovation-Adaptive Homotopy Scheduling}
The filtered confidence score is mapped online to the homotopy parameter. A generic exponential map is
\begin{equation}
\epsilon_k = \epsilon_{\min} + (\epsilon_{\max}-\epsilon_{\min})\left(1-e^{-\beta \bar{s}_k}\right),
\label{eq:eps_map_basic}
\end{equation}
with $\beta>0$ a sensitivity coefficient. In the implementation used for the numerical study, this mapping may also include a deadband and saturation to prevent small nominal innovation fluctuations do not trigger unnecessary smoothing. Regardless of the exact tuning, the interpretation is the same when the estimator is confident, $\epsilon_k$ remains near the aggressive low value, and the controller behaves close to the fuel-optimal bang-bang regime. When the innovation and MTF activity increase, $\epsilon_k$ rises, and the control law is automatically relaxed toward a smoother and more conservative regime.

\begin{figure}[H]
    \centering
    \includegraphics[width=\linewidth]{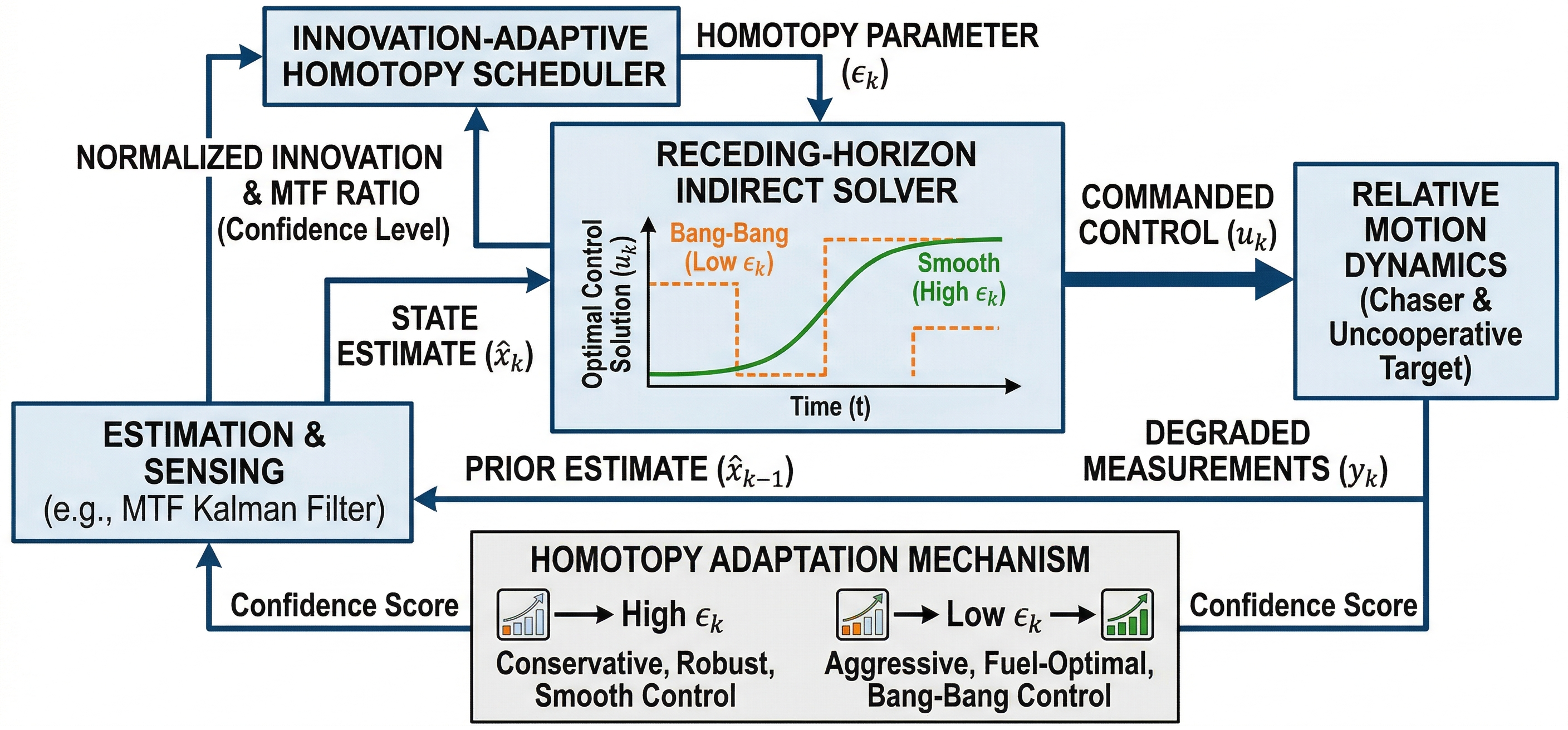}
    \caption{System-level schematic of the proposed tightly-coupled adaptive homotopy estimation and guidance architecture.}
    \label{fig:sch}
\end{figure}

The proposed coupling creates two distinct robustness mechanisms. The first is the estimator-side mechanism. The MTF term inflates the effective measurement covariance when anomalous measurements are detected, preventing the filter from overreacting to corrupted data. The second is the guidance-side mechanism. The innovation-based homotopy scheduling changes the numerical structure of the indirect optimal control problem itself, making the commanded trajectory less aggressive when state confidence degrades. Together, these two loops produce a tightly-coupled estimation and guidance architecture in which both the state estimate and the control regularization respond to the same confidence information. Figure~\ref{fig:sch} summarizes the logic of the proposed tightly-coupled architecture.

\section{Numerical Results and Discussion}
\label{sec:results}
The terminal rendezvous case considered in this study is not an abstract, standalone benchmark, but is derived from the proximity operations environment developed in our prior vision-based relative navigation work. In that earlier effort, a tumbling target spacecraft was modeled in Blender and observed by a virtual camera during a fly-around sequence, and the relative pose was estimated using an adaptive unscented Kalman filtering framework [\citen{candan, candanMdpi}]. The present work builds on that same environment by treating the resulting relative geometry and terminal approach segment as the guidance and control testbed. The scenario is motivated by autonomous rendezvous with large uncooperative spacecraft and debris objects, with particular emphasis on the ENVISAT target problem that has become a representative benchmark for active debris removal and on-orbit servicing studies [\citen{ESTABLE202052, envisatdl, cakal, envisat1998}]. For the current study, the relative motion is modeled in the Clohessy-Wiltshire frame, and the chaser is initialized at
\[
\mathbf{x}_0 =
[0.03031809,\ 0,\ 31.16639,\ -0.02963377,\ 0.04570523,\ 0]^T,
\]
with position expressed in kilometers and velocity in kilometers per second. The terminal target state is taken as the origin, corresponding to rendezvous with the target-centered frame over an \(800\) s horizon. The control input is bounded by a defined maximum acceleration, and open-loop energy-optimal and fuel-optimal solutions are first computed as reference trajectories. Then, three closed-loop architectures were compared under the same measurement degradation. A plain Kalman filter with fixed bang-bang control, an MTF-based Kalman filter with fixed bang-bang control, and the proposed MTF-based filter coupled with innovation-adaptive homotopy scheduling. For reference, the open-loop energy-optimal and fuel-optimal trajectories were also computed offline. Moreover, for reproducibility, Table~\ref{tab:hyperparams} summarizes the principal simulation and scheduler hyperparameters used in the closed-loop study.

\begin{table}[t]
\centering
\caption{Simulation and adaptive-scheduler hyperparameters used in the numerical study.}
\label{tab:hyperparams}
\begin{tabular}{l l l}
\hline
\textbf{Parameter} & \textbf{Value} & \textbf{Description} \\
\hline
$T_f$ & $800$ s & Terminal rendezvous horizon \\
$\Delta t$ & $1$ s & Simulation / filter update step \\
$u_{\max}$ & $8\times10^{-4}$ km/s$^2$ & Maximum control acceleration \\
$\beta$ & $0.25$ & Innovation-to-homotopy sensitivity \\
$m_{\mathrm{meas}}$ & $3$ & Measurement dimension \\
$\rho_{\mathrm{score}}$ & $0.85$ & Low-pass filter coefficient for score \\
$k_{\mathrm{MTF}}$ & $12.0$ & Weight on MTF severity in composite score \\
$\mathrm{score}_{\max}$ & $50.0$ & Saturation limit for composite score \\
$\epsilon_{\min}$ & $0.0$ & Minimum homotopy parameter \\
$\epsilon_{\max}$ & $1.0$ & Maximum homotopy parameter \\
$\alpha_{\epsilon}$ & $0.10$ & Homotopy smoothing/blending factor \\
$t_{\mathrm{resolve}}$ & $10$ s & Receding-horizon re-solve interval \\
$t_{\mathrm{min,rem}}$ & $20$ s & Minimum remaining horizon for re-solve \\
$R_{\mathrm{nominal}}$ & $\mathrm{diag}(10^{-6},10^{-6},10^{-6})$ km$^2$ & Nominal measurement covariance \\
$Q_{\mathrm{proc}}$ & $\mathrm{diag}(10^{-12},10^{-12},10^{-12},10^{-14},10^{-14},10^{-14})$ & Process covariance \\
$P_0$ & $\mathrm{diag}(10^{-6},10^{-6},10^{-6},10^{-10},10^{-10},10^{-10})$ & Initial state covariance \\
\hline
\end{tabular}
\end{table}

The open-loop baselines achieved
\[
\Delta v_{\mathrm{energy}} = 1.502386\times10^{-1}\ \mathrm{km/s},
\qquad
\Delta v_{\mathrm{fuel}} = 1.128000\times10^{-1}\ \mathrm{km/s},
\]
with negligible terminal miss. In the closed loop, however, the two fixed-$\epsilon$ controllers both failed to satisfy the terminal objective under degraded sensing. Specifically, the plain-KF and MTF-KF bang-bang controllers each produced a terminal miss of approximately
\[
1.60\times10^{-1}\ \mathrm{km},
\]
despite maintaining a control effort slightly below the open-loop fuel-optimal benchmark.

By contrast, the proposed MTF-adaptive homotopy controller reduced the terminal miss to
\[
9.277792\times10^{-4}\ \mathrm{km},
\]
while using
\[
1.437594\times10^{-1}\ \mathrm{km/s}
\]
of cumulative control effort, corresponding to a fuel penalty of \(27.446\%\) relative to the open-loop fuel-optimal solution. Thus, the proposed method traded a moderate increase in control effort for an approximately two-order-of-magnitude improvement in terminal accuracy. In practical terms, the final miss was reduced from about \(160\)~m to less than \(1\)~m.

\begin{figure}
    \centering
    \includegraphics[width=\linewidth]{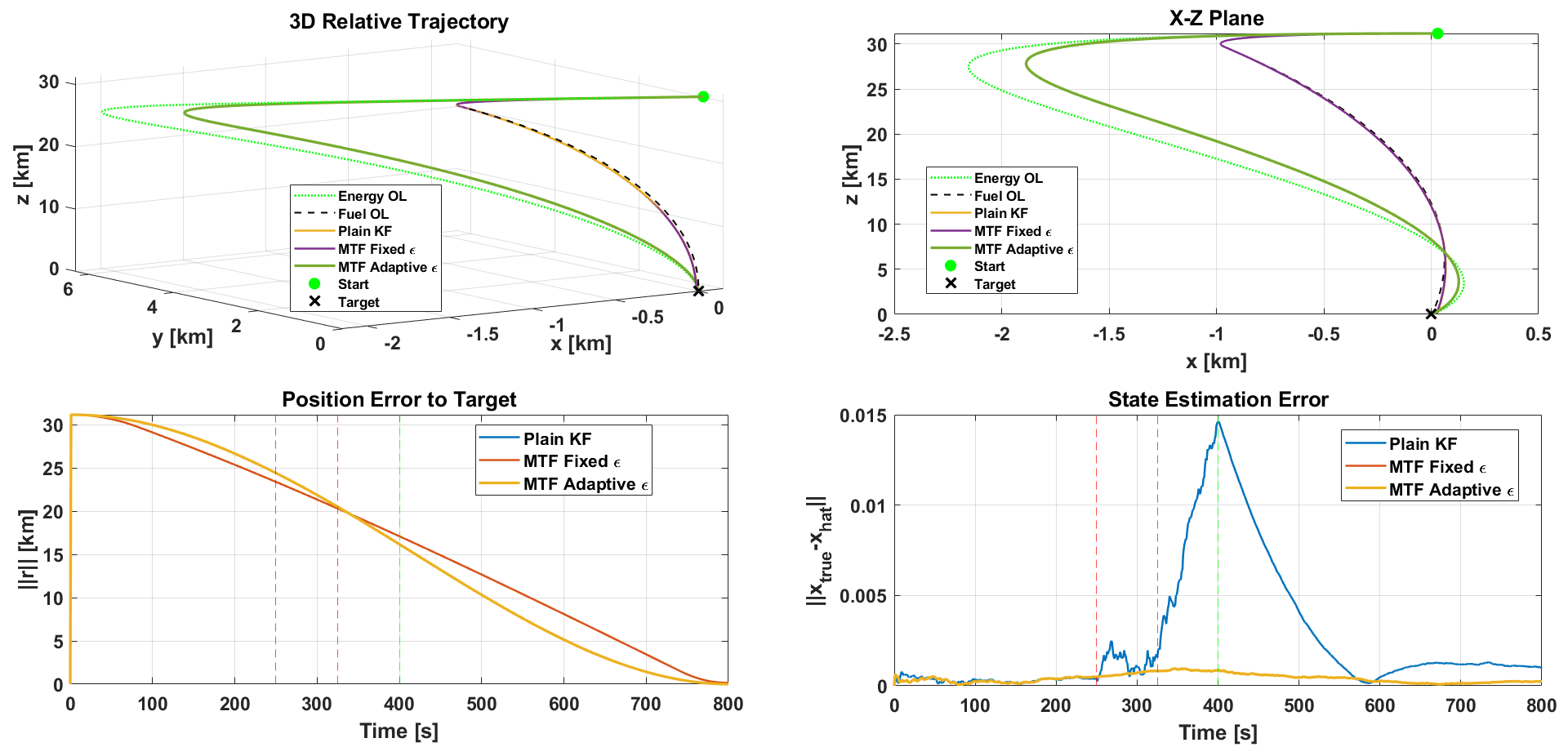}
    \caption{Trajectory and state-level comparison of the open-loop and closed-loop rendezvous solutions. The 3D and \(x\)-\(z\) views show that the proposed MTF-adaptive homotopy controller remains close to the desired terminal approach while the fixed-\(\epsilon\) architectures exhibit greater sensitivity to degraded sensing. The lower panels show the corresponding position error to the target and the state-estimation error, illustrating the improved robustness and recovery behavior of the adaptive architecture.}
    \label{fig:1}
\end{figure}

\begin{figure}
    \centering
    \includegraphics[width=\linewidth]{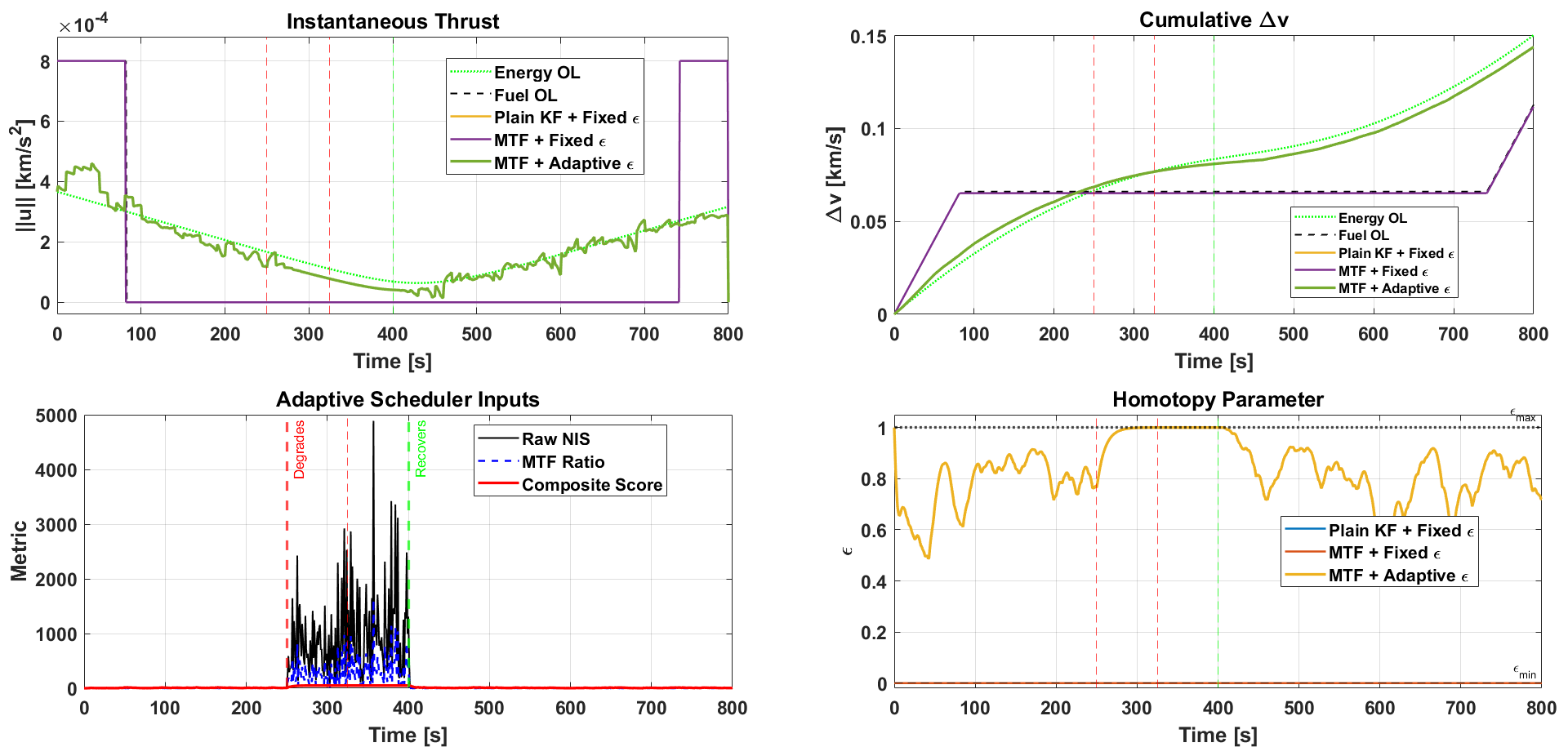}
\caption{Control effort and adaptation behavior for the closed-loop rendezvous case. The upper panels compare the instantaneous thrust magnitude and cumulative \(\Delta v\), showing the robustness versus fuel trade introduced by the adaptive controller. The lower panels show the innovation-based scheduler inputs and the resulting homotopy parameter history, demonstrating how increased innovation activity during sensor degradation drives the controller toward a smoother and more conservative regime.}
    \label{fig:2}
\end{figure}

The visual results in Figs.~\ref{fig:1} and \ref{fig:2} are consistent with the quantitative trends discussed above. Figure~\ref{fig:1} compares the resulting relative trajectories and the corresponding position and estimation errors. In the 3D and \(x\)-\(z\) views, the proposed MTF-adaptive homotopy trajectory remains close to the reference fuel-optimal path while preserving a smoother and more robust terminal approach under degraded sensing. The position error history further shows that, although all controllers begin from the same initial condition, only the adaptive architecture maintains reliable convergence to the target during the degradation interval. The estimation error plot also illustrates that the MTF-based filtering framework suppresses the growth of state estimation error during the imposed anomaly window, with the adaptive case exhibiting the smallest sustained error and the cleanest recovery behavior. 

Figure~\ref{fig:2} highlights the corresponding control-level and adaptation-level behavior. The instantaneous thrust and cumulative-\(\Delta v\) plots show that the fixed-\(\epsilon\) controllers retain a more brittle bang-bang structure, whereas the adaptive controller deliberately departs from the purely fuel-optimal policy and expends additional control effort in exchange for substantially improved terminal robustness. The scheduler input plot clarifies the mechanism behind this behavior: once the innovation statistics increase during the degradation interval, the composite confidence metric rises sharply, triggering an increase in the homotopy parameter. This behavior is shown explicitly in the homotopy parameter history, where \(\epsilon\) is elevated during periods of degraded confidence and then relaxes as the sensing condition recovers. Together, these figures confirm that the proposed method does not merely improve estimation in isolation, but actively reshapes the closed-loop guidance law in response to estimator confidence. 

\begin{figure}
    \centering
    \includegraphics[width=\linewidth]{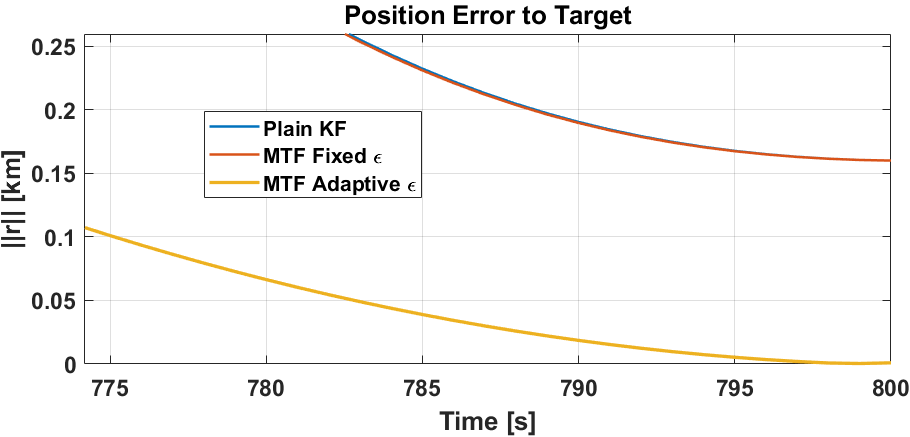}
    \caption{Close-up view of the terminal position error near the end of the rendezvous.}
    \label{fig:3}
\end{figure}

\begin{figure}
    \centering
    \includegraphics[width=\linewidth]{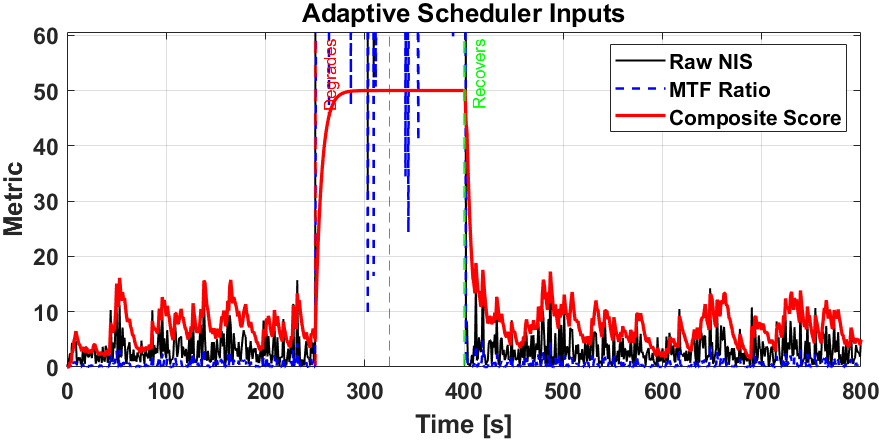}
    \caption{Close-up view of the innovation-based scheduler inputs during the sensor degradation and recovery interval.}
    \label{fig:4}
\end{figure}

Figure~\ref{fig:3} focuses on the final portion of the position error history and makes the terminal performance gap especially clear. While the fixed \(\epsilon\) architectures flatten out at a nonzero miss distance, the MTF-adaptive homotopy controller continues reducing the terminal error and converges much closer to the target. Figure~\ref{fig:4} zooms in on the scheduler inputs during the degradation and recovery window, showing more clearly how the raw innovation, MTF activity, and composite score evolve together. Close-up views reinforce the interpretation that the proposed framework improves both the internal adaptation logic and the final rendezvous accuracy.

\subsection{Role of MTF and Adaptive Homotopy}
An important observation is that the plain-KF and MTF-KF fixed bang-bang controllers were nearly indistinguishable. This indicates that, for the present benchmark, measurement covariance inflation alone was insufficient to alter the closed-loop guidance outcome in a meaningful way. Although the MTF mechanism improved the estimator's robustness to suspicious measurements, the controller itself remained fixed in the aggressive bang-bang regime, so the overall trajectory remained brittle. To isolate the role of MTF within the adaptive architecture, an additional comparison study was performed in which the homotopy scheduler remained active but the MTF contribution was disabled. In that case, the plain adaptive controller still succeeded qualitatively, achieving a terminal miss of
\[
1.417456\times10^{-2}\ \mathrm{km},
\]
but at a substantially higher control cost of
\[
2.053365\times10^{-1}\ \mathrm{km/s},
\]
corresponding to an \(82.036\%\) fuel penalty relative to the open-loop fuel-optimal baseline. When MTF was restored within the same adaptive architecture, the terminal miss was reduced to
\[
9.277792\times10^{-4}\ \mathrm{km},
\]
while the cumulative control effort decreased to
\[
1.437594\times10^{-1}\ \mathrm{km/s},
\]
corresponding to a reduced fuel penalty of \(27.446\%\). Thus, adding MTF to the adaptive controller reduced the terminal miss from approximately \(14.2\)~m to \(0.93\)~m while also lowering the control effort by roughly \(30\%\).

These results clarify the relative roles of the two mechanisms. First, adaptive homotopy is the dominant source of robustness, since the controller remains qualitatively successful even when the MTF term is removed. Second, MTF provides a meaningful secondary enhancement by sharpening the confidence signal used for scheduling and by reducing unnecessary conservatism within the adaptive guidance loop. In other words, MTF is not the sole reason the adaptive controller works, but it makes the adaptive architecture both more accurate and more efficient.

\begin{figure}[H]
    \centering
    \includegraphics[width=\linewidth]{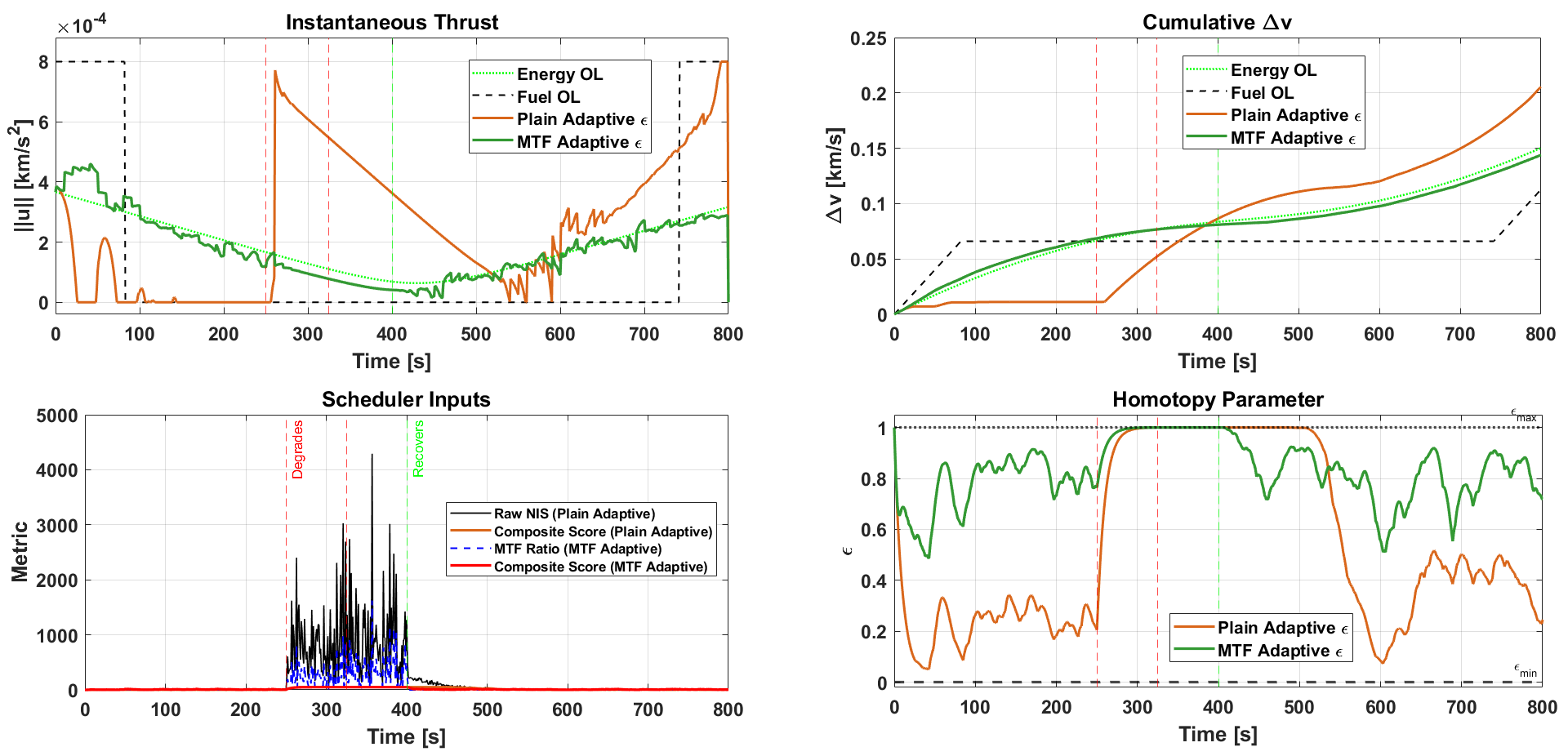}
    \caption{Adaptive-only comparison comparing the plain adaptive-homotopy controller and the MTF-adaptive controller.}
    \label{fig:5}
\end{figure}

\begin{figure}[H]
    \centering
    \includegraphics[width=\linewidth]{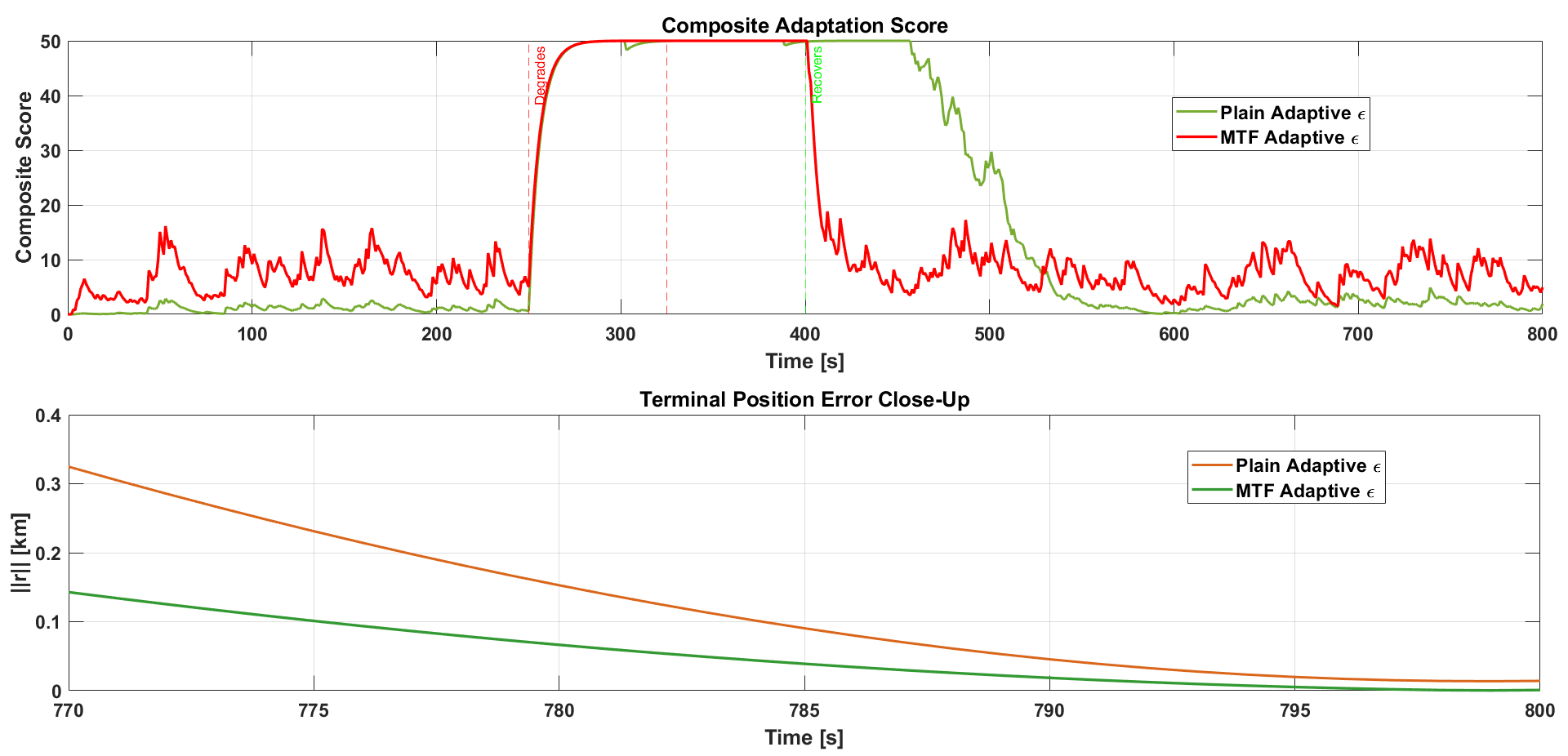}
    \caption{Close-up view of the adaptive-only comparison study.}
    \label{fig:6}
\end{figure}

The comparison study also reveals an important computational advantage. The plain adaptive controller required a mean re-solve time of \(324.18\)~ms, a maximum re-solve time of \(2210.89\)~ms, and achieved a solve success rate of only \(67.95\%\). By contrast, the MTF-adaptive controller achieved a mean solve time of \(54.89\)~ms, a maximum solve time of \(163.61\)~ms, and a solve success rate of \(100\%\). Therefore, in addition to improving terminal accuracy and reducing control expenditure, the MTF-enhanced adaptive architecture also made the indirect receding-horizon solve substantially more reliable and more compatible with online execution. Figures~\ref{fig:5} and \ref{fig:6} provide a visual interpretation of this comparative result. Figure~\ref{fig:5} compares the control effort, cumulative \(\Delta v\), scheduler inputs, and homotopy histories for the two adaptive controllers. The plain adaptive case reacts strongly to the degradation interval and remains significantly more conservative even after recovery, whereas the MTF-adaptive case exhibits a more structured confidence response and a less wasteful control history. Figure~\ref{fig:6} further emphasizes this distinction through close-up views of the composite score and the terminal position error. In particular, the zoomed terminal-error plot makes clear that adaptive homotopy alone is sufficient to recover the rendezvous qualitatively, but the addition of MTF drives the solution to a much tighter final convergence.

Taken together, these results support the main claims of the paper. The central benefit comes from tightly coupling estimator confidence to the homotopy regularization of the indirect solver. The dominant robustness gain arises from allowing the controller itself to become less aggressive when confidence deteriorates, while the MTF mechanism strengthens this coupling by refining the innovation statistics, reducing sensitivity to axis-specific anomalous measurements, and improving the computational reliability of the adaptive closed-loop guidance process.

\subsection{Code and Media Availability}
Rendered Blender videos of the terminal rendezvous scenario are provided as supplementary media and are available together with the implementation repository.
\begin{itemize}
    \item Code repository: \href{https://github.com/dukynuke/AAS_AIAA_2026_Code}{GitHub repository}.
    \item Supplementary rendezvous video: \href{https://youtu.be/TcvcICAxUGo}{Unlisted video link}.
\end{itemize}

\section{Conclusion}
\label{sec:conclusion}
This paper presented a tightly-coupled estimation and guidance architecture for robust low-thrust rendezvous in which estimator confidence is used not only to improve the state estimate, but also to reshape the numerical structure of the onboard indirect optimal control problem. The proposed framework combines a receding-horizon indirect low-thrust guidance law in the Clohessy--Wiltshire frame with a linear Kalman filter employing measurement-side MTF covariance inflation. A composite innovation metric, formed from the normalized innovation and the MTF activity, is mapped online to the homotopy continuation parameter so that the controller becomes less aggressive when sensing degrades and returns toward a fuel-efficient bang-bang regime as confidence recovers.

The numerical study showed that fixed bang-bang guidance remains brittle under degraded sensing, while the proposed MTF-adaptive homotopy controller achieves substantially tighter terminal convergence with only a moderate increase in control effort. The adaptive-only comparison study further clarified that innovation-adaptive homotopy is the dominant robustness mechanism, whereas MTF acts as a secondary enhancement that improves accuracy, efficiency, and numerical reliability within the adaptive architecture. In this sense, the main benefit of the proposed method is not simply improved filtering, but the closed-loop coupling that allows estimator confidence to directly regularize the guidance law online. The computational results also support the practical promise of the method. For the demonstrated linearized rendezvous case, the proposed receding-horizon implementation achieved consistently fast and reliable solve times that are compatible with the available update interval, while the MTF update itself contributed negligible overhead relative to the shooting-based re-optimization.

Overall, the results show that robust low-thrust close-proximity operations benefit from breaking the usual practical separation between estimation confidence and guidance aggressiveness. Rather than blindly enforcing an aggressive fuel-optimal solution under degraded sensing, the proposed framework enables a continuous and physically interpretable trade between fuel optimality and robustness. Future work will extend the method beyond the linear CW setting to higher-fidelity nonlinear relative dynamics, richer measurement models, and hardware-targeted timing studies, and will further investigate scenarios in which MTF plays an even stronger role under structured axis-dependent outliers and model mismatch. A particularly important next step is to embed explicit keep-out and no-go regions into the adaptive guidance problem, including exclusion cones generated by the tumbling motion of solar panels and other appendages of derelict spacecraft during close-proximity operations.

\newpage

\bibliographystyle{AAS_publication}   
\bibliography{references}   

\end{document}